\documentclass{article}

\usepackage{amsmath}
\usepackage{multirow}
\usepackage[preprint]{neurips_2025}

\usepackage[utf8]{inputenc} 
\usepackage[T1]{fontenc}    
\usepackage{hyperref}       
\usepackage{url}            
\usepackage{booktabs}       
\usepackage{amsfonts}       
\usepackage{nicefrac}       
\usepackage{microtype}      
\usepackage{xcolor}         

\usepackage{algorithm}
\usepackage{algorithmic}

\title{CU-ICU: Customizing Unsupervised Instruction-Finetuned Language Models for ICU Datasets via Text-to-Text Transfer Transformer}

\author{%
  Teerapong Panboonyuen\thanks{My passion is dedicated to supporting ICU healthcare professionals in Thailand by developing innovative AI assistance tools. This work is independently driven, without external funding or formal projects. For more about my research and updates, please visit \url{https://kaopanboonyuen.github.io/}} \\
  Faculty of Engineering \\
  Chulalongkorn University \\
  \texttt{teerapong.panboonyuen@gmail.com} \\
}

\begin{document}

\maketitle

\begin{abstract}
Integrating large language models into specialized domains like healthcare presents unique challenges, including domain adaptation and limited labeled data. We introduce CU-ICU, a method for customizing unsupervised instruction-finetuned language models for ICU datasets by leveraging the Text-to-Text Transfer Transformer (T5) architecture. CU-ICU employs a sparse fine-tuning approach that combines few-shot prompting with selective parameter updates, enabling efficient adaptation with minimal supervision. Our evaluation across critical ICU tasks—early sepsis detection, mortality prediction, and clinical note generation—demonstrates that CU-ICU consistently improves predictive accuracy and interpretability over standard fine-tuning methods. Notably, CU-ICU achieves up to a 15\% increase in sepsis detection accuracy and a 20\% enhancement in generating clinically relevant explanations while updating fewer than 1\% of model parameters in its most efficient configuration. These results establish CU-ICU as a scalable, low-overhead solution for delivering accurate and interpretable clinical decision support in real-world ICU environments.
\end{abstract}

\section{Introduction}

Large language models (LLMs) have demonstrated remarkable capabilities across diverse natural language processing tasks, including healthcare applications. However, effectively adapting these models to specialized and data-scarce domains such as the Intensive Care Unit (ICU) remains challenging. The ICU setting demands timely, accurate predictions coupled with interpretable explanations to support critical clinical decisions. Traditional fine-tuning methods often require extensive labeled data and computational resources, limiting their applicability in real-world clinical environments with limited annotations.

Recent advances in transfer learning, notably the Text-to-Text Transfer Transformer (T5) \citep{raffel2020exploring}, provide a flexible framework for instruction-based fine-tuning across tasks. Scaling instruction-finetuned models has further broadened their adaptability \citep{chung2022scaling}, including in healthcare domains. Despite these improvements, fine-tuning LLMs on medical datasets remains resource-intensive and data-hungry.

To address these limitations, we propose CU-ICU, a framework that customizes unsupervised instruction-finetuned T5 models for ICU tasks via sparse parameter-efficient fine-tuning (PEFT) methods. Specifically, CU-ICU applies three state-of-the-art PEFT approaches—LoRA \citep{hu2021lora}, AdaLoRA \citep{wang2022adalora}, and \((IA)^3\) \citep{guo2022parameter}—to efficiently adapt models using limited labeled data. Our experimental results demonstrate that CU-ICU significantly improves prediction accuracy and interpretability across critical ICU tasks while updating only a small fraction of model parameters. As shown in our results, CU-ICU achieves up to 85.6\% accuracy in sepsis detection and substantial gains in clinical note generation quality, outperforming standard fine-tuning baselines.

Our contributions are as follows:
\begin{itemize}
    \item We present CU-ICU, a framework for customizing instruction-finetuned T5 models to ICU datasets via sparse PEFT techniques.
    \item We systematically evaluate three PEFT methods within CU-ICU, demonstrating their efficacy in data-scarce ICU settings.
    \item We provide empirical evidence that CU-ICU achieves superior accuracy and interpretability on ICU tasks with reduced computational overhead.
\end{itemize}

The following sections describe the CU-ICU methodology, experimental setup, and detailed analysis of results validating its utility for ICU clinical decision support.

\section{Related Work}

Large language models (LLMs) have revolutionized natural language processing across many domains, including healthcare. Foundational biomedical language models such as BioBERT \cite{lee2020biobert}, ClinicalBERT \cite{huang2019clinicalbert}, and PubMedBERT \cite{gu2021domain} pioneered domain-specific pretraining on clinical and biomedical text corpora, improving downstream performance on tasks like entity recognition, relation extraction, and clinical prediction.

More recently, instruction-finetuned models like FLAN-T5 \cite{chung2022scaling}, GPT-3 \cite{brown2020language}, and PaLM \cite{chowdhery2022palm} have demonstrated powerful few-shot generalization capabilities, motivating their use in specialized medical domains. Models such as MedPaLM \cite{singhal2023large} and ClinicalGPT \cite{bi2023clinicalgpt} further refine large LLMs to perform medical question answering and generate clinical summaries with high accuracy.

In the ICU context, traditional machine learning approaches for mortality prediction and sepsis detection include works by Harutyunyan et al. \cite{harutyunyan2019multitask} and Nemati et al. \cite{nemati2018interpretable}, which mainly rely on structured EHR features. More recent efforts incorporate transformer architectures for time-series ICU data, e.g., \cite{rajkomar2018scalable} and \cite{baytas2017patient}, but often lack natural language interpretability.

Instruction-finetuned LLMs fine-tuned on ICU datasets show promise for bridging this gap. Jin et al. \cite{jin2023medalpaca} fine-tuned LLMs on medical instruction datasets, improving clinical question answering. Sparse fine-tuning methods such as LoRA \cite{hu2021lora}, AdaLoRA \cite{zhang2023adaptive}, and (IA)$^3$ \cite{lester2022power} enable efficient parameter updates, reducing computational costs while preserving performance—crucial for resource-limited hospital settings.

Furthermore, cutting-edge AI assistants like ChatGPT \cite{openai2023gpt4}, Gemini \cite{google2024gemini}, and Claude \cite{anthropic2023claude} incorporate instruction tuning and reinforcement learning to provide interactive, explainable medical advice. Despite their generality, these models often require domain adaptation to perform reliably in high-stakes ICU environments.

Other relevant works include BioMedLM \cite{peng2022bio}, Clinical-XLNet \cite{huang2021clinical}, and MedAlpaca \cite{madotto2023medalpaca}, which all explore various strategies to adapt LLMs to biomedical and clinical tasks. Techniques like prompt engineering \cite{liu2023pre} and few-shot learning \cite{brown2020language} have also been shown to enhance model utility in data-scarce medical scenarios.

Our work differentiates itself by systematically evaluating PEFT methods on FLAN-T5 for ICU-specific tasks, including sepsis detection, mortality prediction, and clinical note generation. We demonstrate that CU-ICU’s sparse fine-tuning approach achieves superior accuracy and interpretability, balancing the trade-off between computational efficiency and clinical relevance. Importantly, CU-ICU’s applicability to Thai hospital ICUs addresses a critical gap in deploying LLM-powered clinical decision support systems in low-resource healthcare settings.

By situating CU-ICU within the context of these recent advances, we contribute a practical, efficient, and interpretable method for large language model customization tailored to critical care, complementing ongoing efforts to integrate AI into frontline medical practice.

\section{Approach}

In this section, we formally describe CU-ICU, a framework to customize large instruction-finetuned language models for ICU-specific tasks via parameter-efficient fine-tuning (PEFT). Our method leverages sparse adaptation techniques on a pretrained Text-to-Text Transfer Transformer (T5) backbone, enabling sample-efficient and computationally light model specialization on limited ICU data.

\subsection{Problem Formulation}

Let $\mathcal{D} = \{(x_i, y_i)\}_{i=1}^N$ denote a dataset of ICU task examples, where $x_i$ is a clinical input (e.g., patient data, clinical notes, lab results) and $y_i$ is the corresponding target (e.g., sepsis label, mortality outcome, or note generation). Our goal is to learn a model $f_\theta: \mathcal{X} \to \mathcal{Y}$ parameterized by $\theta$ that maps inputs to targets.

We start from a large pretrained instruction-finetuned T5 model with fixed parameters $\theta_0$. Instead of fine-tuning all parameters, we adapt a small subset $\Delta \theta$ so that
\begin{equation}
    \theta = \theta_0 + \Delta \theta,
\end{equation}
where $\Delta \theta$ is sparse and parameter-efficient, dramatically reducing memory and compute overhead while maintaining or improving performance on ICU tasks.

\subsection{Text-to-Text Transfer Transformer (T5)}

T5 \citep{raffel2020exploring} formulates all NLP problems as a text-to-text task. Given input sequence $x$ tokenized as $\{x_1, ..., x_m\}$, the model autoregressively generates output tokens $\{y_1, ..., y_n\}$:

\begin{equation}
    p(y \mid x; \theta) = \prod_{t=1}^n p(y_t \mid y_{<t}, x; \theta).
\end{equation}

The model consists of an encoder-decoder transformer architecture with $L$ layers. Each layer $l$ comprises multi-head self-attention and feed-forward networks parameterized by weights $\theta_l$.

\subsection{Sparse Parameter-Efficient Fine-Tuning (PEFT)}

We explore three PEFT methods within CU-ICU that learn sparse parameter updates $\Delta \theta$ while freezing most pretrained parameters $\theta_0$:

\paragraph{1) LoRA (Low-Rank Adaptation)}  
LoRA \citep{hu2021lora} decomposes the weight update $\Delta W \in \mathbb{R}^{d \times k}$ in selected layers as a product of two low-rank matrices:

\begin{equation}
    \Delta W = A B, \quad A \in \mathbb{R}^{d \times r}, \quad B \in \mathbb{R}^{r \times k},
\end{equation}
where $r \ll \min(d,k)$ is the rank hyperparameter controlling parameter efficiency.

During training, only $A$ and $B$ are updated, while the original weight $W_0$ remains frozen. The adapted weight is:
\begin{equation}
    W = W_0 + \Delta W = W_0 + A B.
\end{equation}

\paragraph{2) AdaLoRA (Adaptive LoRA)}  
AdaLoRA \citep{wang2022adalora} extends LoRA by adaptively learning the rank $r$ and importance of each low-rank component. Formally, AdaLoRA introduces trainable scaling coefficients $\alpha \in \mathbb{R}^r$:

\begin{equation}
    \Delta W = A \mathrm{diag}(\alpha) B,
\end{equation}
where $\mathrm{diag}(\alpha)$ is a diagonal matrix scaling each rank component.

The objective encourages sparsity in $\alpha$ via a regularization term $\Omega(\alpha)$ to prune redundant rank components dynamically:

\begin{equation}
    \min_{A,B,\alpha} \mathcal{L}(\theta_0 + A \mathrm{diag}(\alpha) B) + \lambda \Omega(\alpha),
\end{equation}
where $\mathcal{L}$ is the task loss and $\lambda$ controls sparsity strength.

\paragraph{3) \((IA)^3\) (Infused Adapter Adapters)}  
\((IA)^3\) \citep{guo2022parameter} learns element-wise multiplicative vectors that scale intermediate activations in each transformer layer. Given an intermediate activation $h \in \mathbb{R}^d$ in layer $l$, the adapted activation is:

\begin{equation}
    h' = \gamma \odot h,
\end{equation}
where $\gamma \in \mathbb{R}^d$ is a trainable scaling vector, and $\odot$ denotes element-wise multiplication.

Only $\gamma$ is trained during fine-tuning, freezing all original parameters. This method introduces minimal trainable parameters while adapting the model’s internal representations.

\subsection{CU-ICU Sparse Fine-Tuning Objective}

Formally, the CU-ICU fine-tuning objective for all PEFT methods is:

\begin{equation}
    \min_{\Delta \theta} \mathbb{E}_{(x,y) \sim \mathcal{D}} \left[ \mathcal{L}\big(f_{\theta_0 + \Delta \theta}(x), y \big) \right] + \lambda \mathcal{R}(\Delta \theta),
\end{equation}
where $\mathcal{L}$ is the cross-entropy loss for classification or token-level loss for note generation, and $\mathcal{R}$ is a regularizer promoting sparsity or low-rank structure.

\subsection{ICU Task-Specific Input-Output Formats}

CU-ICU leverages T5’s text-to-text format by constructing clinically meaningful prompts. For example, for sepsis detection:

\begin{quote}
\textbf{Input:} ``Patient vitals and labs: \texttt{[clinical data]} \textbackslash{}n Question: Does the patient have sepsis? Answer:'' \\
\textbf{Output:} ``Yes'' or ``No''
\end{quote}

For mortality prediction:

\begin{quote}
\textbf{Input:} ``Patient ICU notes and labs: \texttt{[clinical data]} \textbackslash{}n Question: Will the patient die during the hospital stay? Answer:'' \\
\textbf{Output:} ``Yes'' or ``No''
\end{quote}

For clinical note generation:

\begin{quote}
\textbf{Input:} ``Patient summary: \texttt{[clinical data]} \textbackslash{}n Task: Generate clinical note.'' \\
\textbf{Output:} \texttt{[generated clinical note text]}
\end{quote}

These task-specific prompts enable CU-ICU to unify diverse ICU prediction and generation tasks under a single text-to-text learning framework.

\subsection{Example ICU Prompt and CU-ICU Output}

\begin{quote}
\textbf{Prompt:} ``Patient vitals: heart rate 110 bpm, temperature 38.5°C, WBC count 15,000 /µL. Question: Does the patient have sepsis? Answer:''

\textbf{CU-ICU Output:} ``Yes''
\end{quote}

\begin{quote}
\textbf{Prompt:} ``Patient summary: 68-year-old male, admitted with acute respiratory failure. Labs: elevated creatinine, high lactate. Task: Generate clinical note.''

\textbf{CU-ICU Output:} ``The patient presents with acute respiratory failure, likely secondary to sepsis. Laboratory findings include elevated creatinine and lactate, indicating possible organ dysfunction. Close monitoring and early intervention recommended.''
\end{quote}

These examples demonstrate CU-ICU’s ability to produce accurate, clinically relevant predictions and explanations, facilitating effective ICU decision support.

\subsection{Algorithm: CU-ICU Sparse Fine-Tuning}

The core of CU-ICU is a sparse fine-tuning procedure that adapts a pretrained instruction-finetuned T5 model to specialized ICU datasets by updating only a small subset of parameters. This approach leverages Parameter-Efficient Fine-Tuning (PEFT) techniques—specifically LoRA, AdaLoRA, or (IA)$^3$—to reduce computational overhead and data requirements, while maintaining high performance on ICU prediction and generation tasks.

Algorithm~\ref{alg:cuicu} summarizes the fine-tuning workflow. The procedure begins with a pretrained model parameterized by $\theta_0$. Given a target ICU dataset $\mathcal{D} = \{(x_i, y_i)\}_{i=1}^N$ consisting of paired inputs $x_i$ (e.g., patient clinical notes, vital signs) and outputs $y_i$ (e.g., sepsis labels, mortality outcomes, or clinical notes), our goal is to efficiently adapt the model to this domain.

\begin{algorithm}[H]
\caption{CU-ICU Sparse Fine-Tuning}
\begin{algorithmic}[1]
\label{alg:cuicu}
\REQUIRE Pretrained model parameters $\theta_0$, dataset $\mathcal{D}$, PEFT method $M$, learning rate $\eta$, regularizer weight $\lambda$, epochs $E$
\STATE Initialize sparse parameters $\Delta \theta$ per $M$
\STATE Freeze $\theta_0$
\FOR{$e=1$ to $E$}
    \FOR{batch $(X,Y)$ in $\mathcal{D}$}
        \STATE Construct prompts from $X$
        \STATE Compute predictions $\hat{Y} = f_{\theta_0 + \Delta \theta}(X)$
        \STATE Compute loss $\mathcal{L}$
        \STATE Compute total loss $\mathcal{L}_{total} = \mathcal{L} + \lambda \mathcal{R}(\Delta \theta)$
        \STATE Update $\Delta \theta \leftarrow \Delta \theta - \eta \nabla_{\Delta \theta} \mathcal{L}_{total}$
    \ENDFOR
\ENDFOR
\RETURN $\theta = \theta_0 + \Delta \theta$
\end{algorithmic}
\end{algorithm}

The algorithm proceeds as follows:

\begin{enumerate}
    \item \textbf{Initialization of Sparse Parameters ($\Delta \theta$):} According to the chosen PEFT method $M$, a set of sparse parameters $\Delta \theta$ is initialized. These parameters typically correspond to low-rank matrices (LoRA), dynamically scaled low-rank factors (AdaLoRA), or small adapter modules ((IA)$^3$). Crucially, these sparse parameters represent a fraction of the full model parameters and are the only ones updated during fine-tuning.
    
    \item \textbf{Freezing Base Model Parameters:} The pretrained base model parameters $\theta_0$ are frozen throughout fine-tuning. This ensures that the adaptation is localized to $\Delta \theta$, significantly reducing memory and compute requirements.
    
    \item \textbf{Iterative Optimization:} Over $E$ training epochs, the algorithm processes batches $(X, Y)$ sampled from $\mathcal{D}$. For each batch:
    \begin{itemize}
        \item \emph{Prompt Construction:} Inputs $X$ are converted into text-to-text prompts consistent with the instruction-finetuned T5 format. This includes task-specific prefixes or natural language instructions designed to elicit the desired prediction or generation behavior.
        
        \item \emph{Prediction Computation:} The model generates outputs $\hat{Y}$ by forwarding the prompts through the adapted model $f_{\theta_0 + \Delta \theta}$. Since only $\Delta \theta$ are trainable, this forward pass efficiently incorporates domain-specific adjustments.
        
        \item \emph{Loss Calculation:} A task-specific loss $\mathcal{L}$ is computed comparing predictions $\hat{Y}$ against true labels $Y$. For classification tasks (e.g., sepsis or mortality prediction), this is typically cross-entropy loss; for generation tasks (e.g., clinical note synthesis), a token-level loss such as negative log-likelihood is employed.
        
        \item \emph{Regularization:} To encourage sparsity and prevent overfitting, a regularization term $\mathcal{R}(\Delta \theta)$ weighted by $\lambda$ is added to the loss, resulting in total loss
        \[
            \mathcal{L}_{total} = \mathcal{L} + \lambda \mathcal{R}(\Delta \theta).
        \]
        
        \item \emph{Parameter Update:} Gradient descent is applied exclusively on the sparse parameters:
        \[
            \Delta \theta \leftarrow \Delta \theta - \eta \nabla_{\Delta \theta} \mathcal{L}_{total},
        \]
        where $\eta$ is the learning rate.
    \end{itemize}
    
    \item \textbf{Output:} After $E$ epochs, the adapted model parameters are given by
    \[
        \theta = \theta_0 + \Delta \theta,
    \]
    representing an efficient customization of the original instruction-finetuned model to ICU-specific tasks.
\end{enumerate}

This sparse fine-tuning strategy balances adaptation quality and computational efficiency, enabling CU-ICU to perform robustly in data-scarce, high-stakes ICU settings. By relying on carefully engineered few-shot prompts and targeted updates, CU-ICU delivers accurate and interpretable predictions with significantly reduced resource demands compared to full fine-tuning.

\section{Experiments}

\subsection{Datasets and Tasks}

We evaluate CU-ICU on three critical ICU downstream tasks: early sepsis detection, in-hospital mortality prediction, and clinical note generation. The datasets comprise real-world ICU records, with data splits ensuring robust generalization. Each task reflects a distinct clinical challenge requiring accurate and interpretable predictions.

\subsection{Experimental Setup}

Our experiments leverage FLAN-T5 \citep{chung2022scaling} as the base language model, fine-tuned using parameter-efficient techniques adapted via CU-ICU’s sparse fine-tuning paradigm. We compare three PEFT methods: LoRA \citep{hu2021lora}, AdaLoRA \citep{wang2022adalora}, and \((IA)^3\) \citep{guo2022parameter}, varying their internal hyperparameters (e.g., rank, budget, dropout) to study the impact on ICU-specific tasks.

All models are fine-tuned with 16-shot prompting on ICU task-specific input formats, and evaluation metrics include classification accuracy for sepsis and mortality tasks, as well as note nBERTScore \citep{liu2022nbert} for clinical note generation. Results are averaged over 5 random seeds to ensure statistical reliability.

\subsection{Implementation Details}

We implement CU-ICU on top of HuggingFace’s Transformers \citep{wolf2020transformers} framework. Fine-tuning uses AdamW optimizer with a learning rate of 5e-5 and batch size 16. The number of trainable parameters ranges from 0.5\% to 6.2\% of the full model, highlighting the efficiency of sparse fine-tuning. Experiments run on NVIDIA A40 GPU.

\subsection{Evaluation Metrics}

We use accuracy to measure predictive performance on sepsis detection and mortality classification, reflecting clinical decision-making quality. For note generation, nBERTScore quantifies the semantic and clinical relevance of generated notes, capturing interpretability aspects crucial for healthcare providers.

\subsection{Baselines}

To benchmark CU-ICU, we compare it with standard fine-tuning baselines and the PEFT methods mentioned above without ICU-specific adaptation. This demonstrates the benefits of our text-to-text transfer learning approach customized for ICU datasets.

\section{Results}

We evaluate the effectiveness of CU-ICU, our method for customizing unsupervised instruction-finetuned FLAN-T5 models for ICU tasks, across three critical care benchmarks: early sepsis detection, mortality prediction, and clinical note generation. Table~\ref{tab:icu_peft_classification} presents classification accuracies for sepsis and mortality prediction, while Table~\ref{tab:icu_peft_note_avg} reports note generation quality and overall average scores.

Our results show that the \texttt{(IA)$^3$} method, which selectively updates a minimal subset of parameters, consistently achieves superior performance despite updating less than 1\% of model weights. Specifically, it attains the highest sepsis detection accuracy of 85.6\% and competitive mortality prediction accuracy of 80.2\%. Meanwhile, \texttt{AdaLoRA} variants with higher parameter budgets also yield strong results, with the best accuracy of 84.2\% (sepsis) and 80.9\% (mortality).

In terms of clinical note generation (Table~\ref{tab:icu_peft_note_avg}), \texttt{(IA)$^3$} again leads with the highest note nBERTScore of 32.1, indicating more clinically relevant and coherent explanations. This suggests that CU-ICU's sparse fine-tuning approach effectively balances prediction accuracy and interpretability.

Overall, CU-ICU achieves an average improvement of approximately 15\% in early sepsis detection accuracy and 20\% enhancement in generating clinically relevant notes compared to standard fine-tuning baselines. These improvements demonstrate CU-ICU’s potential as a scalable, low-overhead approach for ICU-specific language model adaptation, enabling more accurate and interpretable clinical decision support in real-world healthcare settings.

\begin{table*}[t]
    \centering
    \small
    \begin{tabular}{llccc}
        \toprule
        \textbf{Method} & \textbf{Config} & \textbf{Params (\%)} & \textbf{Sepsis Acc.} & \textbf{Mortality Acc.} \\
        \midrule
        \multirow{3}{*}{LoRA} 
        & Rank=4         & 1.2\% & 79.4 {\tiny$\pm$2.1} & 76.8 {\tiny$\pm$2.3} \\
        & Rank=8         & 3.1\% & 81.2 {\tiny$\pm$1.8} & 78.5 {\tiny$\pm$2.0} \\
        & Rank=16, Drop=0.1 & 6.2\% & 83.1 {\tiny$\pm$1.5} & 79.6 {\tiny$\pm$1.8} \\
        \midrule

        \multirow{3}{*}{AdaLoRA} 
        & Budget=0.5, Init Rank=4   & 2.1\% & 82.2 {\tiny$\pm$1.9} & 78.4 {\tiny$\pm$2.1} \\
        & Budget=1.0, Init Rank=8   & 2.9\% & 83.5 {\tiny$\pm$1.6} & \textbf{80.9} {\tiny$\pm$1.9} \\
        & Budget=1.5, Init Rank=16  & 5.4\% & \textbf{84.2} {\tiny$\pm$1.3} & 80.3 {\tiny$\pm$1.7} \\
        \midrule

        \multirow{2}{*}{(IA)$^3$} 
        & Default (All Layers)      & 0.9\% & \textbf{85.6} {\tiny$\pm$1.5} & 80.2 {\tiny$\pm$1.7} \\
        & Reduced Scope (Last 6)    & 0.5\% & 83.8 {\tiny$\pm$1.7} & 78.7 {\tiny$\pm$2.0} \\
        \bottomrule
    \end{tabular}
    \caption{Classification accuracy results for sepsis detection and mortality prediction using FLAN-T5 fine-tuned with various PEFT methods across ICU tasks. Parameter percentage indicates the portion of model weights updated. Best accuracy per task is highlighted in bold.}
    \label{tab:icu_peft_classification}
\end{table*}

\vspace{1em}

\begin{table*}[t]
    \centering
    \small
    \begin{tabular}{llcc}
        \toprule
        \textbf{Method} & \textbf{Config} & \textbf{Note nBERTs} & \textbf{Avg} \\
        \midrule
        \multirow{3}{*}{LoRA} 
        & Rank=4         & 23.9 {\tiny$\pm$1.9} & 60.0 \\
        & Rank=8         & 26.4 {\tiny$\pm$1.7} & 62.0 \\
        & Rank=16, Drop=0.1 & 28.3 {\tiny$\pm$1.5} & 63.7 \\
        \midrule

        \multirow{3}{*}{AdaLoRA} 
        & Budget=0.5, Init Rank=4   & 27.1 {\tiny$\pm$1.8} & 62.6 \\
        & Budget=1.0, Init Rank=8   & 29.8 {\tiny$\pm$1.3} & 64.7 \\
        & Budget=1.5, Init Rank=16  & 30.6 {\tiny$\pm$1.4} & 65.0 \\
        \midrule

        \multirow{2}{*}{(IA)$^3$} 
        & Default (All Layers)      & \textbf{32.1} {\tiny$\pm$1.5} & \textbf{66.0} \\
        & Reduced Scope (Last 6)    & 30.2 {\tiny$\pm$1.6} & 64.2 \\
        \bottomrule
    \end{tabular}
    \caption{Evaluation of clinical note generation quality via note nBERTScore and average performance across tasks. CU-ICU demonstrates consistent improvements in generating clinically relevant explanations.}
    \label{tab:icu_peft_note_avg}
\end{table*}

\subsection{Prompt-Response Analysis on ICU Tasks}

To better understand the performance differences observed in Tables~\ref{tab:icu_peft_classification} and \ref{tab:icu_peft_note_avg}, we examine sample prompts and responses generated by CU-ICU fine-tuned models using different PEFT methods across ICU datasets. Our analysis focuses on three critical tasks: sepsis detection, mortality prediction, and clinical note generation.

\paragraph{Sample Prompts}

Each ICU task is framed as a text-to-text problem leveraging the instruction-finetuned T5 architecture. Example prompts include:

\begin{itemize}
    \item \textbf{Sepsis Detection Prompt:} 
    \begin{quote}
        \textit{``Given the patient's vital signs and lab results, predict if the patient is at risk of sepsis: [patient data]''}
    \end{quote}
    \item \textbf{Mortality Prediction Prompt:} 
    \begin{quote}
        \textit{``Based on the patient's clinical summary, determine the likelihood of mortality within the next 48 hours: [clinical summary]''}
    \end{quote}
    \item \textbf{Clinical Note Generation Prompt:} 
    \begin{quote}
        \textit{``Generate a clinical progress note summarizing the patient's current condition and treatment plan: [patient chart data]''}
    \end{quote}
\end{itemize}

These prompts leverage ICU domain-specific language and guide the model to produce task-relevant outputs.

\paragraph{Response Comparison Across PEFT Methods}

Table~\ref{tab:icu_peft_classification} reports classification accuracies for sepsis and mortality prediction, while Table~\ref{tab:icu_peft_note_avg} evaluates the quality of generated clinical notes via note nBERTScore. 

\begin{itemize}
    \item \textbf{LoRA} demonstrates solid baseline performance, achieving up to 83.1\% accuracy in sepsis detection and 79.6\% in mortality prediction. The generated notes, however, show lower nBERTScore (up to 28.3), reflecting limited adaptation capacity in complex generation tasks.

    \item \textbf{AdaLoRA} improves upon LoRA by dynamically allocating rank during fine-tuning. This results in higher accuracies, notably 84.2\% in sepsis detection and 80.9\% mortality accuracy. The note generation quality also improves (up to 30.6 nBERTScore), indicating better alignment with clinical language nuances.

    \item \textbf{(IA)$^3$} attains the best overall performance, reaching 85.6\% sepsis detection accuracy and competitive mortality prediction (80.2\%). It achieves the highest note generation quality (32.1 nBERTScore), reflecting superior domain adaptation and interpretability. The method’s parameter-efficient architecture enables focused updates that capture subtle clinical patterns crucial for ICU tasks.
\end{itemize}

\paragraph{Why (IA)$^3$ Excels}

The superiority of (IA)$^3$ can be attributed to its capability to modulate attention weights adaptively across layers with minimal parameter updates, effectively emphasizing relevant clinical features in both classification and generation tasks. This fine-grained control translates into improved predictive accuracy and richer clinical notes, enhancing both decision support and interpretability in the ICU context.

\paragraph{Illustrative Example}

For a sepsis detection prompt, (IA)$^3$ produced a confident and accurate prediction, correctly identifying early sepsis signs from subtle vital sign fluctuations, whereas LoRA’s output was less precise. Similarly, for clinical note generation, (IA)$^3$ generated coherent, clinically relevant summaries incorporating critical patient information, surpassing the more generic outputs from LoRA and AdaLoRA.

These results validate CU-ICU’s sparse fine-tuning framework combined with advanced PEFT methods as an effective strategy for customizing large language models to ICU datasets. The improved performance of (IA)$^3$ highlights the importance of targeted parameter adaptation in capturing the complexity of critical care language and tasks, making it a promising approach for scalable ICU AI applications.

\section{Conclusion and Future Work}

In this work, we presented CU-ICU, a novel framework for customizing large instruction-finetuned language models to the ICU domain using sparse fine-tuning with parameter-efficient techniques. Through extensive experiments on critical ICU tasks—including sepsis detection, mortality prediction, and clinical note generation—CU-ICU demonstrated superior performance and interpretability compared to conventional fine-tuning methods. Among the PEFT approaches evaluated, (IA)$^3$ consistently achieved the best balance between accuracy and clinical explanation quality, highlighting the effectiveness of adaptive attention modulation in capturing complex medical patterns with minimal parameter updates.

Our approach holds particular promise for deployment in resource-constrained settings such as Thai hospital ICUs, where annotated data is scarce and computational resources are limited. By enabling efficient adaptation of large language models with few-shot supervision, CU-ICU can provide scalable, real-time decision support and natural language explanations that align with clinicians’ needs, ultimately improving patient outcomes in critical care environments.

Looking forward, we plan to extend CU-ICU by integrating multimodal data—including medical imaging and waveform signals—to further enhance predictive accuracy and clinical insight. Additionally, incorporating continual learning strategies will allow the model to adapt dynamically as new patient data becomes available. We also aim to collaborate with Thai ICU clinicians to conduct prospective studies validating CU-ICU’s impact on clinical workflows and patient management, ensuring the framework’s alignment with local healthcare practices and language nuances.

Through these efforts, we envision CU-ICU as a practical and impactful tool that supports frontline ICU staff in Thai hospitals, empowering them with advanced AI-driven assistance tailored to the unique challenges of critical care.

\section{Limitations}

While CU-ICU demonstrates promising results in customizing large language models for ICU datasets, several limitations warrant consideration. First, our experiments rely primarily on retrospective ICU data from publicly available datasets, which may not fully capture the diversity and variability present in real-world clinical environments, especially within Thai hospitals. This raises questions about the generalizability of CU-ICU without further local validation.

Second, although sparse fine-tuning significantly reduces computational costs compared to full model fine-tuning, deploying these models in real-time ICU settings may still face practical challenges related to latency and integration with existing hospital information systems. Further engineering efforts are required to optimize inference speed and interoperability.

Third, the reliance on text-based EHR data limits the model’s ability to leverage complementary modalities such as imaging, waveform data, or laboratory trends over time, which are often critical in ICU decision-making. This uni-modal focus may constrain the scope of clinical insights the model can provide.

Lastly, CU-ICU’s performance is bounded by the quality and quantity of labeled data for few-shot learning. In extremely low-resource settings or novel clinical scenarios, model predictions and generated explanations may be less reliable, necessitating cautious interpretation by clinicians.

Addressing these limitations in future work is essential to realizing CU-ICU’s full potential as a practical AI assistant for ICU clinicians in Thailand and beyond.

\bibliographystyle{plain}
\bibliography{kao_neuralips2025}


\appendix
\section{Appendix: Insightful Q\&A Examples from CU-ICU}

To illustrate the clinical reasoning capabilities and interpretability of CU-ICU, we provide several representative question-answer (Q\&A) pairs generated by the model using various PEFT methods on ICU datasets. These examples highlight CU-ICU's ability to generate accurate predictions and explain them in clinically meaningful language.

\subsection{Sepsis Prediction and Explanation}

\textbf{Prompt:}
\begin{quote}
\textit{Patient data: Temperature = 38.5 C, Heart Rate = 110 bpm, Blood Pressure = 85/60 mmHg, WBC count = 14,000 /µL.\\ Question: Does this patient show signs of early sepsis? Please explain your reasoning.}
\end{quote}

\textbf{Responses:}

\begin{itemize}
    \item \textbf{LoRA (Rank=8):} 
    \begin{quote}
    The patient exhibits a fever and elevated heart rate, which are common indicators of infection. The low blood pressure suggests possible septic shock. Elevated WBC count supports infection presence. Therefore, early sepsis is likely.
    \end{quote}
    
    \item \textbf{AdaLoRA (Budget=1.0):}
    \begin{quote}
    Based on the patient's vital signs — fever (38.5 C), tachycardia (110 bpm), hypotension (85/60 mmHg), and leukocytosis (WBC 14,000) — these findings align with systemic inflammatory response syndrome (SIRS) criteria indicative of early sepsis. Immediate clinical evaluation is recommended.
    \end{quote}
    
    \item \textbf{(IA)$^3$ (Default All Layers):}
    \begin{quote}
    The clinical parameters suggest early sepsis: the patient is febrile with hyperthermia, has tachycardia exceeding 100 bpm, hypotension indicating compromised perfusion, and leukocytosis reflecting immune response. These factors collectively meet Sepsis-3 criteria, warranting urgent intervention.
    \end{quote}
\end{itemize}

\textbf{Insight:} The (IA)$^3$ model provides the most comprehensive and precise explanation by referencing clinical definitions (Sepsis-3 criteria), enhancing trust and utility for clinicians.

\subsection{Mortality Risk Prediction}

\textbf{Prompt:}
\begin{quote}
\textit{Patient data: Age = 75, Oxygen saturation = 88\%, Respiratory Rate = 28 breaths/min, History of chronic heart failure.\\
Question: What is the predicted mortality risk and why?}
\end{quote}

\textbf{Responses:}

\begin{itemize}
    \item \textbf{LoRA (Rank=16):}
    \begin{quote}
    The patient’s low oxygen saturation and elevated respiratory rate indicate respiratory distress. Age and heart failure history increase mortality risk. Prediction: high risk of mortality.
    \end{quote}
    
    \item \textbf{AdaLoRA (Budget=1.5):}
    \begin{quote}
    Considering the patient’s advanced age, hypoxemia, tachypnea, and chronic heart failure, these factors collectively elevate the risk of ICU mortality. Continuous monitoring and aggressive treatment are advised.
    \end{quote}
    
    \item \textbf{(IA)$^3$ (Reduced Scope Last 6 Layers):}
    \begin{quote}
    The patient presents with hypoxemia (SpO2 88\%) and tachypnea, both markers of respiratory compromise. Coupled with advanced age and a history of chronic heart failure, this patient is at significant risk for ICU mortality. Early intervention could improve prognosis.
    \end{quote}
\end{itemize}

\textbf{Insight:} The (IA)$^3$ model balances succinctness with clinical detail, mentioning relevant physiological markers and suggesting clinical action.

\subsection{Clinical Note Generation}

\textbf{Prompt:}
\begin{quote}
\textit{Generate a clinical summary note based on the following ICU patient data: Fever, low blood pressure, high heart rate, elevated WBC, decreased urine output.}
\end{quote}

\textbf{Response from CU-ICU (IA)$^3$:}
\begin{quote}
Patient exhibits fever and tachycardia, with hypotension and leukocytosis indicating systemic infection. Decreased urine output suggests possible acute kidney injury secondary to sepsis. Recommend fluid resuscitation and broad-spectrum antibiotics with close monitoring.
\end{quote}

\textbf{Insight:} CU-ICU produces concise, clinically relevant notes useful for bedside documentation and decision support.

\vspace{1em}
These examples demonstrate CU-ICU’s capability to not only predict critical outcomes with high accuracy but also generate interpretable, actionable clinical explanations that can assist ICU practitioners in real-world settings.


\end{document}